\documentclass[conference]{IEEEtran}
\IEEEoverridecommandlockouts

 \usepackage[numbers,sort&compress]{natbib}
\usepackage{amsmath,amssymb,amsfonts}

\usepackage{booktabs}
\usepackage{array}
\usepackage{tabularx}

\usepackage{amsmath}
\usepackage{algorithm}
\usepackage{algpseudocode}
\usepackage{graphicx}
\usepackage{textcomp}
\usepackage{xcolor}
\usepackage{tabularx}

\usepackage{subcaption} 

\def\BibTeX{{\rm B\kern-.05em{\sc i\kern-.025em b}\kern-.08em
    T\kern-.1667em\lower.7ex\hbox{E}\kern-.125emX}}
\begin{document}

\title{Better by Comparison: Retrieval-Augmented Contrastive Reasoning for Automatic Prompt Optimization\\
}


\author{\IEEEauthorblockN{Juhyeon Lee*\thanks{The authors denoted as * have contributed equally to this work. Author order is determined alphabetically by first name.}}
\IEEEauthorblockA{Peking University \\
Beijing, China \\
leejuhyeon@pku.edu.cn}
\and
\IEEEauthorblockN{Wonduk Seo*}
\IEEEauthorblockA{Enhans \\
Seoul, South Korea \\
wonduk@enhans.ai}
\and
\IEEEauthorblockN{Hyunjin An}
\IEEEauthorblockA{Enhans \\
Seoul, South Korea \\
hyunjin@enhans.ai}
\and
\IEEEauthorblockN{Seunghyun Lee}
\IEEEauthorblockA{Enhans \\
Seoul, South Korea \\
seunghyun@enhans.ai}
\and
\IEEEauthorblockN{Yi Bu†\thanks{† denotes corresponding author.}}
\IEEEauthorblockA{Peking University \\
Beijing, China \\
buyi@pku.edu.cn}
}


\maketitle

\begin{abstract}
Automatic prompt optimization has recently emerged as a strategy for improving the quality of prompts used in Large Language Models (LLMs), with the goal of generating more accurate and useful responses. However, most prior work focuses on direct prompt refinement or model fine-tuning, overlooking the potential of leveraging LLMs' inherent reasoning capability to learn from contrasting examples. In this paper, we present \textbf{Contrastive Reasoning Prompt Optimization (CRPO)}, a novel framework that formulates prompt optimization as a retrieval-augmented reasoning process. Our approach retrieves top-$k$ reference prompt-response pairs from the \emph{HelpSteer2} dataset, an open-source collection where each response is annotated for helpfulness, correctness, coherence, complexity, and verbosity, and constructs two complementary optimization paradigms: (1) \emph{tiered contrastive reasoning}, where the LLM compares high-, medium-, and low-quality exemplars (both prompts and responses) to refine its own generation through reflective reasoning, and (2) \emph{multi-metric contrastive reasoning}, where the LLM analyzes the best exemplars along each evaluation dimension and integrates their strengths into an optimized prompt. By explicitly contrasting high- and low-quality exemplars, CRPO enables the model to deduce why certain prompts succeed while others fail, thereby achieving more robust and interpretable optimization. Experimental results on the HelpSteer2 benchmark demonstrate that CRPO significantly outperforms baselines. Our findings highlight the promise of contrastive, retrieval-augmented reasoning for advancing automatic prompt optimization.
\end{abstract}

\begin{IEEEkeywords}
Automatic Prompt Optimization, Contrastive Reasoning, Retrieval-Augmented Generation, Large Language Models (LLMs), Helpfulness Alignment.
\end{IEEEkeywords}

\begin{figure*}
    \centering
    \includegraphics[width=\linewidth]{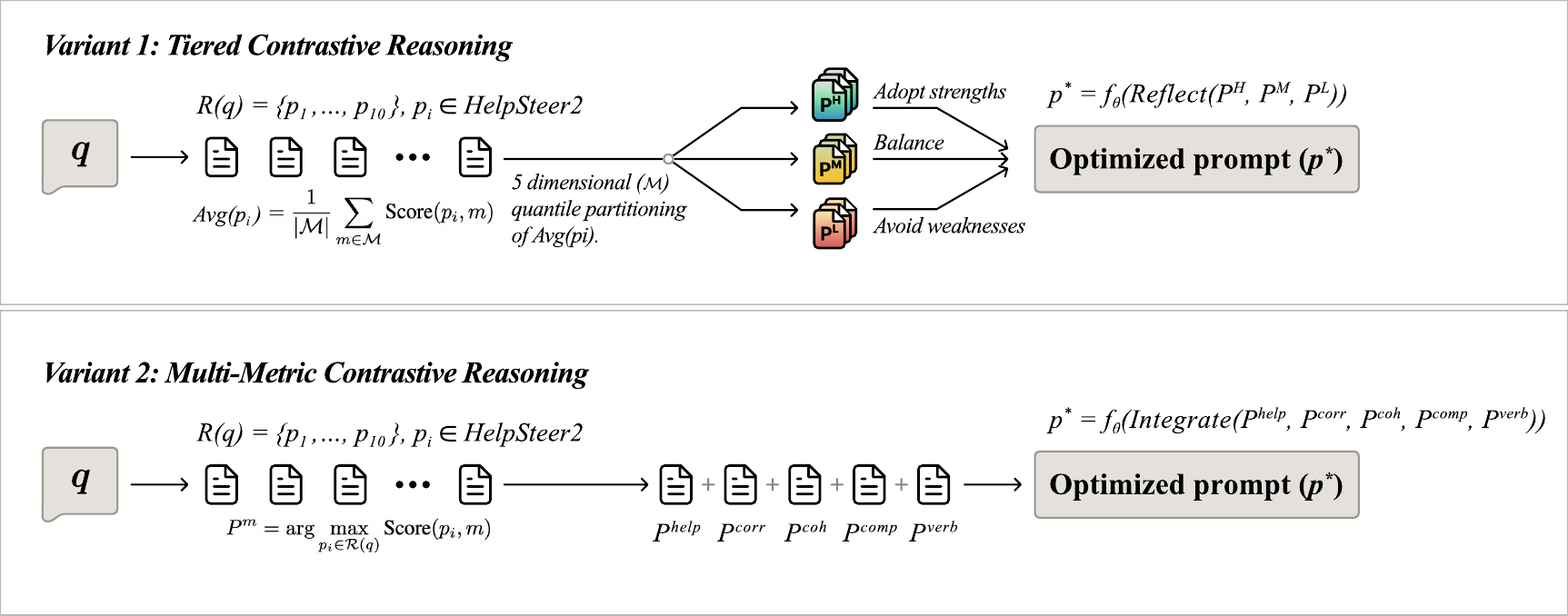}
    \caption{\textbf{Overview of CRPO}: The framework of CRPO-\textit{Tiered Contrastive Reasoning(top)} and CRPO-\textit{Multi-Metric Contrastive Reasoning(bottom).}}
    \label{fig:framework}
\end{figure*}

\section{Introduction}
Large Language Models (LLMs) have demonstrated remarkable capabilities across a wide spectrum of natural language processing (NLP) tasks, including reasoning, summarization, and code generation. However, their performance is highly sensitive to the quality of the input prompt \cite{zhao2021calibrate,lu2021fantastically}. As a result, automatic refining prompts to elicit better responses has become a critical research direction.

Early prompt optimization methods focused on continuous soft prompt tuning \cite{lester2021power,li2021prefix,hu2022lora} or discrete token search \cite{shin2020autoprompt,deng2022rlprompt,zhang2022tempera}, which were effective in white-box settings but inapplicable to black-box LLMs accessed via API. More recent work has shifted to black-box approaches that explore prompts through trial-and-error, such as PromptAgent’s Monte Carlo Tree Search \cite{wang2023promptagent}, hierarchical multi-agent workflows \cite{liu2024towards}, and multi-agent topology optimization \cite{zhou2025multi}. In parallel, some methods treat LLMs themselves as optimizers, for instance, OPRO \cite{yang2023large} frames optimization as iterative natural language refinement, while evolutionary approaches like PromptBreeder \cite{fernando2023promptbreeder} and branching structures such as AMPO \cite{yang2024ampo} adapt prompts through mutation, selection, and pruning.

Despite these advances, current methods share several key limitations: (1) They often optimize prompts in isolation, failing to learn from the comparative lessons offered by better and worse exemplars; (2) Many approaches also depend on handcrafted optimization pipelines, reducing their generality and scalability across domains; (3) Finally, most existing methods focus primarily on improving answer quality, often neglecting human-centered dimensions such as interpretability and usability, which are essential for practical deployment in real-world human–AI interaction.

To address these limitations, we introduce \textbf{Contrastive Reasoning Prompt Optimization (CRPO)}, a retrieval-augmented framework that explicitly leverages contrastive reasoning across prompts of varying quality. Unlike methods that only consider prompt text and scores, CRPO retrieves top-$k$ reference prompt-response pairs from the HelpSteer2 dataset~\cite{wang2024helpsteer} using sparse retrieval over prompt text, where each pair is annotated for helpfulness, correctness, coherence, complexity, and verbosity. By incorporating both the retrieved prompts and their responses, CRPO allows the model to reason not only about how prompts are phrased but also how they shape the quality of generated outputs. It then performs two novel optimization strategies: (1) \emph{tiered contrastive reasoning}, where the model reflects on high-, medium-, and low-quality exemplars to refine its own generation, and (2) \emph{multi-metric contrastive reasoning}, where the best exemplars along individual metrics are analyzed and integrated into an optimized prompt.

CRPO improves prompt quality without model updates by explicitly reasoning about why prompts succeed or fail; on HelpSteer2 it consistently outperforms direct generation, Chain-of-Thought prompting (CoT) \cite{wei2022chain}, and simple retrieval augmentation\footnote{Detailed implementations of baselines are detailed in Section IV.}, underscoring the promise of contrastive, retrieval-augmented optimization for aligning LLM outputs toward more helpful, factual, and coherent responses.

\section{Dataset}
We conduct our experiments on the \textbf{HelpSteer2} dataset~\cite{wang2024helpsteer}, a benchmark designed for evaluating prompt optimization and response helpfulness in Large Language Models (LLMs). The dataset consists of human-annotated prompt-response pairs and is divided into training and validation splits. Specifically, it contains a \textbf{training set of 20.3k rows} and a \textbf{validation set of 1.04k rows}, which we adopt for retrieval and evaluation, respectively. 


Each prompt--response pair in HelpSteer2 is annotated across five evaluation dimensions. Scores for each attribute range from \textbf{0 (lowest)} to \textbf{4 (highest)}, providing a fine-grained scale of response quality. These annotations capture diverse aspects of quality, making the dataset suitable for studying contrastive reasoning in prompt optimization.

\begin{table}[htbp]
\centering
\begin{tabular}{@{}l p{7cm}@{}}
\toprule
\textbf{Attribute} & \textbf{Description} \\
\midrule
Helpfulness & Overall helpfulness of the response to the prompt. \\
Correctness & Inclusion of all pertinent facts without errors. \\
Coherence & Consistency and clarity of expression. \\
Complexity & Intellectual depth required to write the response (e.g., basic competency vs.\ domain expertise). \\
Verbosity & Amount of detail included in the response relative to what is asked. \\
\bottomrule
\end{tabular}
\caption{\textbf{HelpSteer2 Annotation Dimensions.} Five human-annotated metrics, each scored on a 0--4 scale.}
\label{tab:dataset_attributes}
\end{table}

\begin{table*}[t]
  \centering
  \small
  \setlength{\tabcolsep}{6pt}
  \renewcommand{\arraystretch}{1.06}
  \begin{tabular}{l c c c c c | c}
    \toprule
    \textbf{Model} & \textbf{Helpfulness} & \textbf{Correctness} & \textbf{Coherence} & \textbf{Complexity} & \textbf{Verbosity} & \textbf{Avg. Score} \\
    \midrule
    \multicolumn{7}{l}{\textit{GPT-4o}} \\
    Direct Generation & 0.3616 & 0.4700 & 0.7723 & 0.3280 & 0.4913 & 0.4846 \\
    Chain-of-Thought (CoT) & 0.2981 & 0.3958 & 0.7169 & 0.2888 & 0.4709 & 0.4341 \\
    Retrieval Augmented Generation (RAG) & 0.4903 & 0.5745 & 0.8642 & 0.4161 & \underline{0.6567} & 0.6003 \\
    CRPO-Tiered Contrastive Reasoning$^{\dagger}$ & \textbf{0.5274} & \underline{0.6006} & \underline{0.8711} & \underline{0.4360} & 0.6506 & \underline{0.6171} \\
    CRPO-Multi-Metric Contrastive Reasoning$^{\dagger}$ & \underline{0.5219} & \textbf{0.6274} & \textbf{0.8876} & \textbf{0.4386} & \textbf{0.6982} & \textbf{0.6347} \\
    \addlinespace[2pt]
    \multicolumn{7}{l}{\textit{LLaMA-3-8B}} \\
    Direct Generation & 0.2616 & 0.3636 & 0.7078 & 0.2942 & 0.4421 & 0.4139 \\
    Chain-of-Thought (CoT) & 0.1600 & 0.2545 & 0.6291 & 0.2421 & 0.3812 & 0.3334 \\
    Retrieval Augmented Generation (RAG) & 0.3990 & 0.4711 & 0.7989 & 0.3722 & 0.5814 & 0.5245 \\
    CRPO-Tiered Contrastive Reasoning$^{\dagger}$ & \underline{0.4224} & \underline{0.5023} & \underline{0.8092} & \underline{0.3993} & \textbf{0.6435} & \underline{0.5554} \\
    CRPO-Multi-Metric Contrastive Reasoning$^{\dagger}$ & \textbf{0.4264} & \textbf{0.5070} & \textbf{0.8159} & \textbf{0.4146} & \underline{0.6404} & \textbf{0.5609} \\
    \bottomrule
  \end{tabular}
  \caption{Comparison across five evaluation metrics—helpfulness, correctness, coherence, complexity, and verbosity—and their mean (Avg. Score). Results are reported separately for GPT-4o and LLaMA-3-8B. The best value within each language model is \textbf{bold}, the second best is \underline{underlined}, and methods marked with † denote our proposed CRPO variants.}
  \label{tab:crpo_metrics_updated}
\end{table*}

\section{Methodology}
We formulate the prompt optimization problem not as directly fine-tuning model parameters, but as a task that maximizes the inherent reasoning capability of LLMs by facilitating learning from reference exemplars. Specifically, our framework \textbf{Contrastive Reasoning Prompt Optimization (CRPO)} first retrieves the top-$k$ relevant prompt-response pairs from the \emph{HelpSteer2} training set, and then applies contrastive reasoning to construct optimized prompts. We design two complementary variants, illustrated in Figure~\ref{fig:framework}.

\subsection{Retrieval of Reference Prompt--Response Pairs}
Given an input query $q$, CRPO retrieves a set of reference prompt--response pairs 
\[
\mathcal{R}(q) = \{(p_1, r_1), \dots, (p_k, r_k)\}, \quad (p_i, r_i) \in \text{HelpSteer2},
\]
where $p_i$ denotes a prompt and $r_i$ its associated response. Each pair is annotated along five evaluation dimensions $\mathcal{M} = \{\text{help}, \text{corr}, \text{coh}, \text{comp}, \text{verb}\}$. To ensure sufficient coverage across all dimensions, we require $k \geq 5$, so that at least one candidate pair is available per metric. These exemplars (including both prompts and responses) serve as contrasting cases that enable the LLM to perform explicit reasoning.

\subsection{Variant 1: Tiered Contrastive Reasoning}
We partition $\mathcal{R}(q)$ into 3 tiers according to overall quality scores. 
Specifically, for each retrieved pair $(p_i, r_i)$, we compute its average score across all five evaluation dimensions:
\begin{equation}
    \text{Avg}(p_i, r_i) = \frac{1}{|\mathcal{M}|} \sum_{m \in \mathcal{M}} \text{Score}((p_i, r_i), m),
\end{equation}
where $\mathcal{M} = \{\text{help}, \text{corr}, \text{coh}, \text{comp}, \text{verb}\}$.
Pairs are then partitioned into high-quality ($P^H$), medium-quality ($P^M$), and low-quality ($P^L$) tiers based on quantile thresholds of $\text{Avg}(p_i, r_i)$. The optimized prompt $p^{*}$ is generated via contrastive reasoning:
\begin{equation}
    p^{*} = f_{\theta}\big(\text{Reflect}(P^{H}, P^{M}, P^{L})\big),
\end{equation}
where $f_{\theta}$ is the LLM and $\text{Reflect}(\cdot)$ instructs the model to (i) avoid weaknesses in $P^L$, (ii) adopt strengths from $P^H$, and (iii) use $P^M$ as a stabilizing anchor that reduces bias.  
In particular, incorporating $P^M$ prevents overfitting to extreme cases, ensuring balanced refinement that maintains robustness while still progressing toward high-quality prompts.

\subsection{Variant 2: Multi-Metric Contrastive Reasoning}
For each metric $m \in \mathcal{M}$, we select the top pair:
\[
P^m = \arg\max_{(p_i, r_i) \in \mathcal{R}(q)} \text{Score}((p_i, r_i), m).
\]
The optimized prompt is then constructed as:
\begin{equation}
    p^{*} = f_{\theta}\big(\text{Integrate}(P^{\text{help}}, P^{\text{corr}}, P^{\text{coh}}, P^{\text{comp}}, P^{\text{verb}})\big),
\end{equation}
where $\text{Integrate}(\cdot)$ encourages the LLM to combine complementary strengths across evaluation axes, grounded in both prompts and their associated responses.

\section{Experiments}

\subsection{Setup}

We evaluate CRPO on the HelpSteer2 dataset, ensuring fairness by using the same retrieval pool and evaluation settings as the baselines. 
\subsubsection{LLM and Retrieval Models}
We use \textit{GPT-4o} and \textit{LLaMA-3-8B}, both with temperature $0$ and otherwise default hyperparameters. BM25 is specifically used for prompt retrieval, selecting top-$k$ reference prompts $\{p_1,\dots,p_k\}$ per query with $k=10$ to balance coverage across the 5 evaluation dimensions and efficiency. 
\subsubsection{Evaluation Model and Metrics}
We employ the \textit{ArmoRM-Llama3-8B-v0.1} reward model as an interpretable multi-objective judge ($>90\%$ benchmark accuracy), which automatically scores response quality.

For evaluation metrics, we assess effectiveness by feeding the optimized prompt $p^{*}$ together with its generated response $r^{*}$ to the evaluation model $\mathcal{E}$, which returns five HelpSteer2 scores—\textit{helpfulness, correctness, coherence, complexity, verbosity}—each in $[0,4]$:
\begin{equation}
    \text{Score}(p^{*}, r^{*}) = \big[ s_{\text{help}}, s_{\text{corr}}, s_{\text{coh}}, s_{\text{comp}}, s_{\text{verb}} \big].
\end{equation}
Each score is normalized by dividing by 4 and then averaged to yield a final $[0,1]$ quality score:
\begin{equation}
    \mathrm{Score}(p^{*}, r^{*})=\tfrac{1}{|\mathcal{M}|}\sum_{m\in\mathcal{M}}\tfrac{s_m(p^{*}, r^{*})}{4}.
\end{equation}

$(p^{*}, r^{*})$ is judged under identical conditions to the baselines.

    

\subsubsection{Baselines}
We compare our framework against three representative baselines, all of which optimize the prompt based only on the prompt text and evaluation scores, without incorporating structured reasoning assets:  
\begin{itemize}
    \item \textbf{Direct Generation}: The LLM directly produces an optimized prompt from the input query, without retrieval or iterative refinement.  
    \item \textbf{Chain-of-Thought (CoT)}: The LLM is instructed to perform step-by-step reasoning on the input query before generating an optimized prompt, relying solely on the original prompt text and scores.  
    \item \textbf{Retrieval-Augmented Generation (RAG)}: The LLM receives the input query along with the top-$k$ retrieved training prompts (with $k=10$ for fair comparison) and generates an optimized prompt based on these examples and their scores, but without leveraging the associated reasoning assets.  
\end{itemize}


\subsection{Main Experiment Results}

Baseline methods demonstrate modest, albeit limited improvements. Direct Generation often produces shallow or inconsistent outputs, since no external context is used. Chain-of-Thought (CoT) prompting encourages stepwise reasoning but tends toward verbosity and repetition without improving factuality. Retrieval-Augmented Generation (RAG) provides contextual grounding through retrieved prompts, but introduces redundancy and lacks a clear mechanism to filter high-quality signals. Importantly, all three baselines optimize prompts only with respect to prompt text and evaluation scores, without leveraging the responses associated with those prompts. As a result, while baselines improve over one another incrementally, they remain insufficient for producing balanced and robust outputs.  

In contrast, CRPO addresses these shortcomings through explicit contrastive reasoning over prompt--response pairs. The multi-metric variant integrates the strongest exemplars along helpfulness, correctness, coherence, complexity, and verbosity, ensuring complementary strengths are preserved. The tiered variant contrasts high-, medium-, and low-quality pairs, adopting strengths from the best, avoiding weaknesses from the worst, and stabilizing with medium-quality anchors. Together, these strategies yield optimized prompts—and consequently generated responses—that are more robust, interpretable, and human-aligned than those produced by Direct Generation, CoT, or RAG.

\subsection{Ablation studies}

\subsubsection{Trimaximal-Prompt Selection (TPS)
}

\begin{figure}[t]
  \centering
  \includegraphics[width=0.8\linewidth]{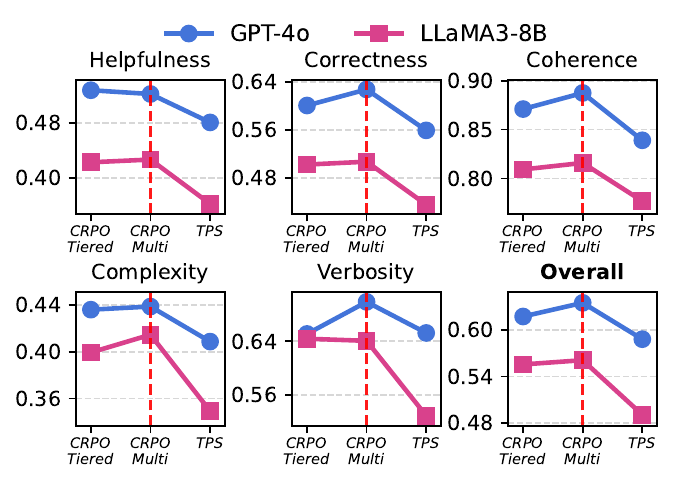}
  \caption{\textbf{Ablation 1.} Performance comparison of the simplified method vs CRPO variants. \textbf{Overall} stands for the average score of 5 metrics.}
  \label{fig:ablation1}
\end{figure}

To assess the role of contrastive reasoning, we conduct an ablation where CRPO is reduced to using only the top-$3$ highest-ranked prompts, as shown in Figure~\ref{fig:ablation1}. In contrast, both CRPO variants consistently outperform the ablated setting by explicitly reasoning over contrasts—adopting strengths from high-quality prompts, avoiding weaknesses from low-quality ones, and integrating complementary aspects across dimensions. This result shows that the gains of CRPO stem not from retrieval alone, but from its reflective reasoning process, which yields more stable and interpretable optimization.

\begin{figure}[t]
  \centering
  \includegraphics[width=0.8\linewidth]{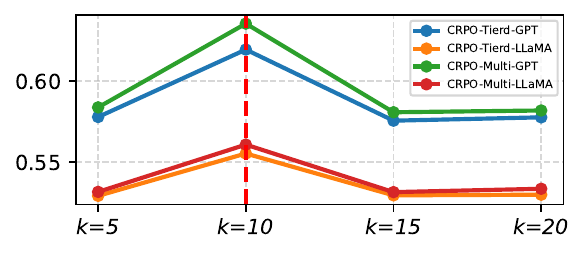}
  \caption{\textbf{Ablation 2.} Overall score comparison across Top-K settings in the RAG stage of CRPO, averaged over five metrics.}
  \label{fig:ablation2}
\end{figure}

\subsubsection{Optimal K validation}

To identify the optimal evidence of retrieved pool size in our RAG framework, we evaluate \textsc{CRPO} with $k \in \{5,10,15,20\}$. As shown in Figure~\ref{fig:ablation2}, performance improves when moving from a small evidence set to a moderate pool, but declines as the pool grows further. Fixing $k=10$ yields the best trade-off between evidence diversity and inference cost, producing the most stable gains across language models and \textsc{CRPO} variants.

\section{Conclusion}

In this paper, we introduced \textbf{Contrastive Reasoning Prompt Optimization (CRPO)}, a retrieval-augmented framework that improves prompt quality through tiered and multi-metric contrastive reasoning. Unlike prior methods that optimize prompts based only on prompt text and scores, CRPO explicitly reflects on high-, medium-, and low-quality prompt--response pairs, leveraging both the prompts and their generated responses to guide optimization. By integrating complementary strengths across multiple evaluation dimensions, CRPO enables LLMs to produce more robust, interpretable, and human-aligned outputs without fine-tuning. Experimental results on the HelpSteer2 dataset show that CRPO consistently outperforms baselines, including direct generation, CoT prompting, and simple retrieval augmentation, with ablation studies confirming the central role of contrastive reasoning over prompt--response pairs in achieving these gains.


\bibliographystyle{ieeetr}
\bibliography{references}

\end{document}